\documentclass[a4paper,conference]{IEEEtran}

\usepackage{graphicx}
\usepackage{amssymb}

\usepackage{tabularx}
\usepackage{booktabs}

\usepackage[colorlinks,citecolor=green,pagecolor=red,pagebackref=true]{hyperref}

\usepackage{amsmath}

\usepackage{cite}

\begin{document}

\title{Tag-Based Attention Guided Bottom-Up Approach for Video Instance Segmentation}

\author{\IEEEauthorblockN{Jyoti Kini and Mubarak Shah}
\IEEEauthorblockA{Center for Research in Computer Vision, University of Central Florida, Orlando, Florida\\
Email: jyoti.kini@knights.ucf.edu, shah@crcv.ucf.edu}}

\maketitle

\begin{abstract}
Video Instance Segmentation is a fundamental computer vision task that deals with segmenting and tracking object instances across a video sequence. Most existing methods typically accomplish this task by employing a multi-stage top-down approach that usually involves separate networks to detect and segment objects in each frame, followed by associating these detections in consecutive frames using a learned tracking head. In this work, however, we introduce a simple end-to-end trainable bottom-up approach to achieve instance mask predictions at the pixel-level granularity, instead of the typical region-proposals-based approach. Unlike contemporary frame-based models, our network pipeline processes an input video clip as a single 3D volume to incorporate temporal information. The central idea of our formulation is to solve the video instance segmentation task as a tag assignment problem, such that generating distinct tag values essentially separates individual object instances across the video sequence (here each tag could be any arbitrary value between 0 and 1). To this end, we propose a novel spatio-temporal tagging loss that allows for sufficient separation of different objects as well as necessary identification of different instances of the same object. Furthermore, we present a tag-based attention module that improves instance tags, while concurrently learning instance propagation within a video. Evaluations demonstrate that our method provides competitive results on YouTube-VIS and DAVIS’19 datasets, and has minimum run-time compared to other state-of-the-art performance methods. 

\end{abstract}

\IEEEpeerreviewmaketitle

\section{Introduction}
In this paper, we tackle one of the emerging video comprehension problems, namely Video Instance Segmentation (VIS), that deals with separating all the object instances with pixel-level labels and establishing association amongst the instances throughout the video sequence. 
While most current approaches divide  
 the overall task into detection, classification, segmentation, and tracking components, we present an integrated end-to-end method of generating consistent discernable tags to distinguish individual object instances across the video clip.

Most contemporary techniques \cite{yang2019video, luiten2019video, Bertasius_2020_CVPR} try to solve the VIS problem using a top-down approach. The top-down methodology is a two-stage disjoint procedure that involves generating dense region proposals and subsequently employing a tracking mechanism to attain instance correspondences across frames. Despite the effective performance, these solutions are restricted by the challenges associated with the design. Firstly, schematics based on multiple disjoint components: object-detector, segmentation, and tracking modules result in a sub-optimal solution due to cumbersome disconnected training. For instance, VIS models that adapt Mask R-CNN \cite{he2017mask} like methods to generate region proposals tend to rely heavily on the ability of the region proposal network to produce well-localized detections. In the case of unstructured videos subject to diverse deformations, extreme lighting conditioning, background clutter, and occlusions, it is likely that that failure of the detector inherently leads to inferior performance of the overall VIS model. Secondly, representation redundancy injected by numerous overlapping region proposals complicates the process of associating objects across frames. Thirdly, frame-based techniques are not equipped to accurately capture  both spatial and temporal  information present in video sequences. 

Our work addresses these concerns by adopting a bottom-up approach that focuses on generating pixel-based feature representations. Primarily, we leverage temporal context by encoding video sequence, instead of frame-based encoding. Additionally, we formulate the VIS task as a tagging problem, where pixels belonging to distinct object instances are assigned different tag values. The solution is based on a  simple formulation, where each distinguishable tag value is assigned to an object instance.  To compel the network to assign the distinct tags to each instance within a video sequence,  we introduce spatio-temporal tagging loss. The proposed loss function constitutes four elements, namely spatial-intra-instance loss, spatial-inter-instance loss, temporal-instance-grouping loss, and temporal-instance-separation loss.

In our approach, encoded video clip embeddings use spatio-temporal attention cues to learn long-range dependencies, while simultaneously capturing local semantic information. The attention guided representation is then passed through a tag generator module to yield well-separated instance tags in all the frames of a video sequence, using the tagging loss. Additionally, the network employs Video Semantic Segmentation (VSS) as a secondary task to inherently improve the primary objective of video instance segmentation. We incorporate a decoder module to generate semantic segmentation masks based on the cross-entropy loss. The input to the decoder is self-attention features and tag-based embeddings. While the self-attention based representation focuses on providing a comprehensive view of the initial input video sequence to the decoder, the tag-based attention module aims at implicitly improving the instance segmentation and learning the association of object instances throughout the video sequence.

In summary, we present an end-to-end trainable VIS approach that does not suffer performance degradation due to  learning conflicts in individual components of  frame-wise object detector and/or tracking. The solution leverages the temporal information by processing video sequences, and gains from the complementary spatio-temporal tagging loss and tag-based attention mechanism. Unlike other state-of-the-art approaches, it is not limited by speed and manages to strike a sound trade-off between performance and run-time. To summarise, this paper makes the  following  contributions:

\begin{itemize}
  \item[$\bullet$] We introduce a competitive bottom-up approach with pixel-level embeddings to solve the VIS task that eliminates training complexities.
  \item[$\bullet$] We employ the temporal context by modeling with video sequence as opposed to contemporary frame-based solutions that either fail to leverage the motion cues well or tend to incorporate computationally intensive elements, such as optical flow.
  \item[$\bullet$] We propose a novel spatio-temporal tagging loss to enable VIS by assigning well-separated tags to each object instance in all frames of a video.
  \item[$\bullet$] We present a tag-based attention module that not only focuses on improving instance tags per frame but also learns propagation of instance masks throughout the video sequence.
  \item[$\bullet$] We, also, generate VSS  as a byproduct, which can be utilized in other independent tasks that demand priors or pseudo labels.
\end{itemize}

\section{Related Work}
\noindent{\bf Video Semantic Segmentation (VSS)}\\
Video semantic segmentation aims at assigning class-aware labels to object pixels, i.e. all the objects belonging to a given category will have a consistent pixel-level label. It is a direct extension of image semantic segmentation task to videos and does not deal with distinguishing object instances or tracking of the objects throughout the clip. In some of the latest works \cite{zhu2017deep, shelhamer2016clockwork, li2018low}, temporal information has also been employed to predict different semantic classes for the pixels across frames.

\noindent{\bf Video Object Segmentation (VOS)}\\
Video object segmentation refers to segmenting a particular object instance through the entire video sequence based on a readily available ground truth mask in the first frame. Most VOS approaches \cite{chen2018blazingly, Zhou_2019_ICCV, Lin_2019_ICCV, oh2018fast, ventura2019rvos, oh2019video, Voigtlaender19CVPR, xu2018youtube, perazzi2017learning, caelles2017one, 8578778}, segment foreground objects in a class-agnostic fashion, unlike popular VIS methods that deal with classifying a predefined set of object categories and then tracking them over the entire sequence.

\noindent{\bf Video Instance Segmentation (VIS)}\\
Video instance segmentation involves segmenting and tracking object instances in videos. Due to the competitive performance rendered by the top-down methodology, most contemporary work on VIS incline towards this approach, despite the inherent training complexities. Pioneering work in VIS by Yang et al. \cite{yang2019video} (MaskTrack R-CNN) adapts the original Mask R-CNN framework and augments it with tracking head to predict object instances and associate instances across frames. MaskProp \cite{Bertasius_2020_CVPR} is an extension to MaskTrack R-CNN with an upgraded mask propagation branch for tracking object instances across the video sequence. Furthermore, the VIS challenge winning approach from ICCV 2019 \cite{luiten2019video} also opts to use the multi-stage approach comprising detection, classification, segmentation, and tracking. Firstly, it leverages the Mask R-CNN detector to generate object region proposals. In the next step, ResNeXt-101 \cite{mahajan2018exploring} classifier is used, followed by UnOVOST \cite{luiten2020unovost} to link segmentation using optical flow. It is only recently that one of the approaches, namely STEm-Seg \cite{Athar_Mahadevan20ECCV}, has made a transition to embrace the bottom-up approach. Although it provides an end-to-end trainable approach, it fails to produce competitive results because of insufficient separation of instance clusters in the embedding space. 

While the approach proposed in this paper is centered around bottom-up fundamentals, we resolve the issues with the pixel-based approach by introducing tag-based attention and spatio-temporal tagging loss. 

\section{Method}
\begin{figure*}
\begin{center}
\includegraphics[width=0.8\textwidth]{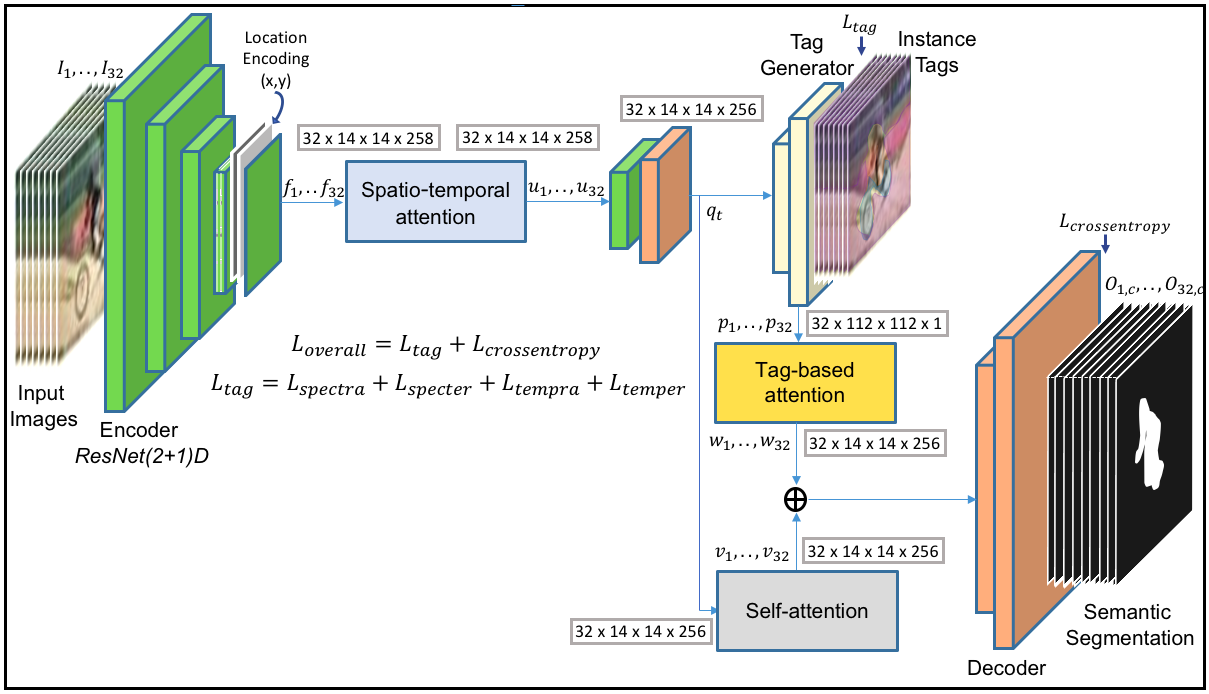}
\end{center}
   \caption{{\bf Network Architecture.} $I_{1}$,..,$I_{32}$ denote input RGB-images passed through ResNet(2+1)D encoder $Enc$. $p_{1}$,..,$p_{32}$ refers to predicted instance tags. $O_{1,c}$,..,$O_{32,c}$ denote binary semantic segmentation masks generated by decoder $Dec$ belonging to $c^{th}$ class category.  $f_{1}$,..,$f_{32}$ and $u_{1}$,..,$u_{32}$ denote input to and output from the spatio-temporal attention module respectively. $q_{t}$ represents features passed to the self-attention module well as tag generator module. $v_{1}$,..,$v_{32}$ correspond to embeddings after the self-attention component. $w_{1}$,..,$w_{32}$ is the resultant output from the tag-based attention module. $\otimes$, $\oplus$, $\odot$ implies  dot product, concatenate operation and element-wise multiplication.} 
\label{fig:Architecture}
\end{figure*}

We propose an end-to-end solution to solve the VIS problem. As illustrated in Figure \ref{fig:Architecture}, the model assigns different tags to each object instance in the input RGB frames of a video clip and additionally generates semantic segmentation as a byproduct. Primarily, an ResNet(2+1)D \cite{tran2018closer} encoder is used to capture the features of the video clip. Thereafter, the resultant features are enriched by passing through spatio-temporal attention module. The generated frame embeddings are then compressed through a bottleneck, which separates the encoder from the tag generator and the decoder. Post the bottleneck component, the network branches into a tag generator that yields instance embeddings and a decoder that provides output semantic segmentation masks. The decoder relies on output from the tag-based attention module and the self-attention module. Though these modules provide input to the decoder, these components play a vital role in implicitly improving the video instance segmentation results through the propagation of loss. While the self-attention module allows us to model long-range dependencies within a frame, the tag-based attention module plays an integral role in improving the instance tags by capturing tag association across frames, which is discussed in detail in Section \ref{Tag-Based_Attention}.

\subsection{Network Architecture}
 In order to provide further insights into the architecture of the model, we first define a set of terminologies. Our network is provided with an input clip $I_{1}$,..,$I_{32}$  comprising $32$ RGB frames to generate instance tags $p_t$  and semantic segmentation masks $O_{t,c}$ belonging to $c$ classes for each frame $t$. We, primarily, use ResNet(2+1)D encoder $Enc$ to generate representation $f_t$. The resultant encoded feature space is further fed to the spatio-temporal attention module to obtain enriched embeddings $u_t$, which capture long-range temporal dependencies, while simultaneously fixating over specific areas in a video frame $t$. These robust internal representations are then synchronously passed through a tag-generator module to generate instance tags $p_t$, and also fed to the self-attention module thereby producing $v_t$. Further, the tag-based attention module improves the instance tags and yields embeddings $w_t$. Thereafter, concatenating $v_t$ with $w_t$ and processing them through the decoder $Dec$ yields semantic segmentation masks $O_{t,c}$ corresponding to $c$ classes.

We use a video sequence of $T$ (in our case, $T=32$) extracted frames and encode it using initial layers till $conv_4$ of ResNet(2+1)D model pre-trained on Kinetics-400. The input dimensions $T \times H \times W$ are therefore downsized to $T \times H' \times W'$ = ${T \times \frac{H}{16} \times \frac{W}{16}}$. Next, we separate these ResNet(2+1)D enriched features $f_t$ across time dimension $f_{t=1,...,32}$, and add 2 additional channels comprising the spatial coordinates to each of these feature embeddings. Thereafter, these individual feature maps $f_{t=1,...,32}$ of size $14 \times 14$ with $256 + 2 = 258$ channels are passed through the spatio-temporal attention module.. The resultant output $u_t$ is further compressed to $32 \times 7 \times 7 \times 512$ in the next layer. Hereafter, we start the up-scaling process to retrieve the appropriately scaled segmentation masks. The first deconvolution layers results in $32 \times 14 \times 14 \times 256$ feature embedding $q_t$, which is further passed through the two-layered tag generator that gives $112 \times 112 \times 1$ sized instances tags $p_t$ for each video frame. In order to improve the generated tags, we introduce tag-based attention module that accepts these generated tags $p_t$ and outputs resultant $14 \times 14  \times 256$ dimension maps $w_t$ per frame. Additionally, the output from the first deconvolution layer, $q_t$, which is $32 \times 14 \times 14 \times 256$, is passed through the self-attention module to yield $32 \times 14 \times 14 \times 256$ dimension embedding $v_t$. We, further, concatenate $v_t$ with $w_t$ and pass through the 3D decoder layers to get the semantic segmentation predictions.

\subsection{Spatio-Temporal Attention}
\begin{figure}[t]
\begin{center}
\includegraphics[width=1\linewidth]{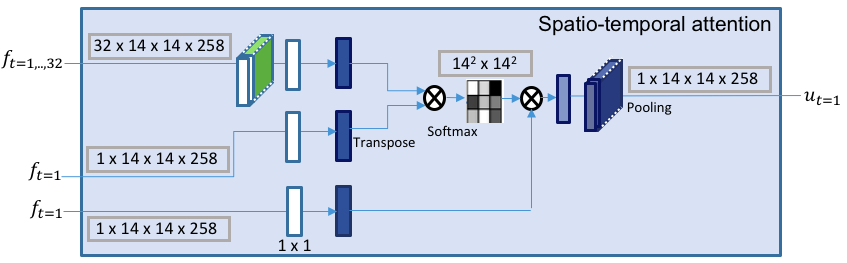}
\end{center}
   \caption{{\bf Spatio-Temporal Attention.} $f_{t=1}$ denotes  feature vector for $1^{st}$  frame, which is input to  to this module . $f_{t=1,...,32}$ represent concatenated input features of  frames 1,...,32\iffalse with an exception of $i^{th}$ time step\fi. $u_{t=1}$ corresponds to output embedding for $1^{st}$ time frame. $\otimes$ denotes inner product}
\label{fig:Spatio_Temporal_Attention}
\end{figure}

We introduce the spatio-temporal attention module to enrich the feature representations by introducing enhanced context associated with spatial priors and long-term motion cues. As shown in Figure \ref {fig:Spatio_Temporal_Attention}, the spatio-temporal attention unit maps the input video feature embedding into query, key, and value. Since our goal is to effectively add context from other frames in the clip to each frame representation, we have a rational choice for the query (Q), key (K), and value (V) tensors:  $f_{t=1,...,32}$ is the query, while $f_{t=1}$ is projected into key and value. Thereafter, we reduce the dimensionality of these components by using $1\times1$ convolutions to manage model complexity while retaining the salient features. Next, we vectorize each component and compute a dot product over the query and the key. The resultant attention map is used for weighted averaging the value. We, further, stack the weighted features based on attention from multiple frames and use pooling across time to get the resultant $u_{t=1}$.

\subsection{Tag-based Attention}
\label{Tag-Based_Attention}
\begin{figure}[t]
\begin{center}
\includegraphics[width=1\linewidth]{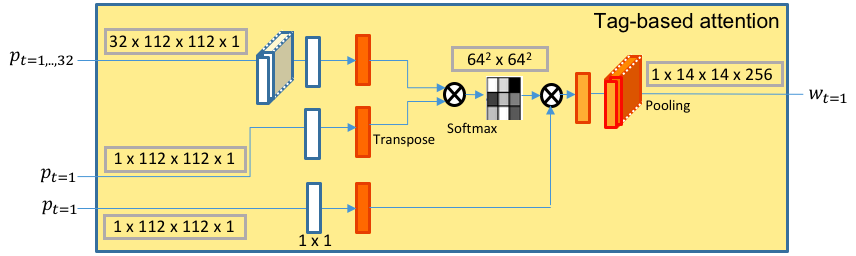}
\end{center}
   \caption{{\bf Tag-Based Attention.} $p_{t=1}$ denotes a tag vector for $1^{st}$ frame, which is input to the module. $p_{t=1,...,32}$ represent concatenated input tags for frames 1,...,32\iffalse with an exception of $i^{th}$ time step\fi. $w_{t=1}$ corresponds to output embedding for $1^{st}$  frame. $\otimes$ denotes inner product.}
\label{fig:Tag_Attention}
\end{figure}

Although most existing competitive VIS approaches opt to solve this problem using a tracking-by-detection paradigm, we adopt a simple pixel-based tag assignment strategy instead. Since assigning distinct tag values to different object instances is the principal theme of our solution, it is essential to improve the separation of instance tags and promote learning related to the association of object instances within a video sequence. The spatio-temporal tagging loss imposes supervision that encourages the allocation of similar tag values to pixels belonging to the same object instance, in a video clip. It also inhibits different object instances from acquiring tag values that do not comply with the necessary margin of instance tag separation. In Section \ref{Training Losses}, we discuss that the loss is not based on drawing a direct comparison with pixels from the ground-truth instance masks, but instead relies on comparison of predicted tag values amongst object instances across the video sequence. To complement this setup, we install the tag-based attention module, so that predicted tags improve owing to the global tag-based context.

\subsection{Self-Attention}
\begin{figure}[t]
\begin{center}
\includegraphics[width=0.8\linewidth]{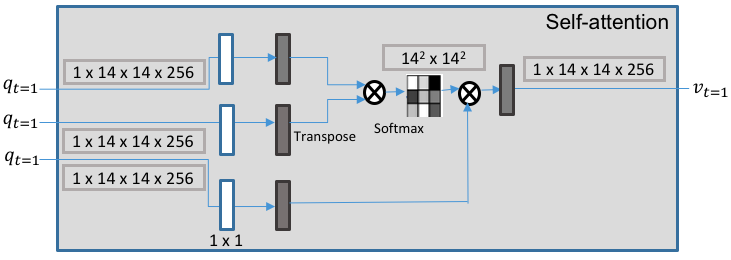}
\end{center}
   \caption{{\bf Self-Attention.} $q_{t=1}$ and $v_{t=1}$ respectively represent input features and output embedding for $1^{st}$ time frame. $\otimes$ denotes inner  product}
\label{fig:Self_Attention}
\end{figure}

Although the primary goal of our architecture is to predict VIS, we adopt a secondary VSS task to support the main objective by implication. It is the propagation of gradients that allows VSS to boost the instance segmentation results consequentially. Introducing the self-attention module plays a vital role in aiding the supporting task of VSS. As illustrated in Figure \ref{fig:Self_Attention}, the self-attention module takes input $q_{t=1}$ and generates enhanced features $v_{t=1}$, where $t=1$ represents the time frame. The stacked representation $v_{t}$ of all the time frames in the given sequence then forms one of the inputs to the semantic segmentation decoder. Here, the embeddings generated by the self-attention module provide an enriched non-local context of input sequence representation to the decoder.

\subsection{Training Losses}
\label{Training Losses}
Based on our network design, we simultaneously generate instance tags as well as semantic segmentation masks. Cross-entropy based loss alone works satisfactorily for semantic segmentation task. However, for tag generator, we propose a novel spatio-temporal tagging loss to place stronger constraints for generating diverse tags amongst the instances. It constitutes four components, namely spatial-intra-instance loss $L_{spectra}$, spatial-inter-instance loss $L_{specter}$, temporal-instance-grouping loss $L_{tempra}$ and temporal-instance-separation loss $L_{temper}$, which allow us to leverage the spatial and temporal context by compelling the network to assign the distinct tags to individual object instances across the video.

Given an object instance $n$, let $m = 1,...M$, where variable $m$ corresponds to a randomly selected pixel associated with the object instance and $M$ is the maximum number of pixels in the instance. Suppose $h_{nm}^{'}$ is the predicted tag value for $m^{th}$ pixel of given instance $n$, then let us define  $ h_{n}$, that is the mean of $M$ tag values for the object instance, as:
\begin{align}
    h_{n} = \frac{1}{M}\sum_{m=1}^{M}{h_{nm}^{'}}.
\end{align}
Next, for $N$ object instances with $M^{'}$ randomly selected pixels of each instance, we formally define spatial-intra-instance loss $L_{spectra}$, that brings all the pixels of a given instance closer, by:
\begin{align}
    L_{spectra} = \frac{1}{N}\sum_{n=1}^{N}\sum_{m=1}^{M^{'}}{(h_{n} - h_{nm}^{'})^{2}}.
\end{align}
In addition to pulling the embeddings of same instance together, we also try to push apart the embeddings corresponding to different instances. We introduce a margin $G$ (in our case, G=3) to provide a permissible threshold on the difference in the embedding space of instance $n$ and $n^{'}$. The  loss $L_{specter}$ is given by:  
\begin{align}
    L_{specter} = \sum_{n^{}=1}^{N-1} \sum_{n^{'}=n+1}^{N} max(0,G-{||h_n - h_{n^{'}}||}).
\end{align}
Furthermore, to incorporate the temporal context associated with each time frame $t$, where $t=1,...T$, we integrate the temporal-instance-grouping loss $L_{tempra}$. It introduces proximity in the given instance across $T$ time-frames in video sequence and is represented as:
\begin{align}
    L_{tempra} = \frac{1}{N}\sum_{n=1}^{N}\sum_{t=1}^{T}{(h_n - h_{nt})^{2}}.
\end{align}
We, also, employ temporal-instance-separation loss $L_{temper}$ to segregate an arbitrary subset of instances $N^{'}$ across time, where $h_{nt}$ denotes mean tag value for the given instance $n$ at time frame $t$:
\begin{align}
    L_{temper} = \sum_{n=1}^{N-1} \sum_{n^{'}=n+1}^{N^{'}} \sum_{t=1}^{T} max(0,G-{||h_{nt} - h_{n^{'}t}||}).
\end{align}
Subsequently, the spatio-temporal tagging loss $L_{tag}$ is defined as:
\begin{align}
    L_{tag} = L_{spectra} + L_{specter} + L_{tempra} + L_{temper}.
\end{align}
Finally, the overall loss is given as follows:
\begin{align}
    L_{overall} = L_{tag} + L_{crossentropy}.
\end{align}

\section{Experimental Setup}
\noindent
\textbf{Datasets}
We perform multiple experiments to evaluate our solution on Youtube-VIS \cite{yang2019video} dataset. Youtube-VIS is a benchmark VIS dataset comprising high-resolution YouTube videos split into 2,238 training sequences, 302 validation sequences, and 343 test sequences. It includes object instances classified into 40 broad categories, such as person, animals, and vehicles. We also perform quantitative analysis on DAVIS'19 Unsupervised Video Object Segmentation \cite{Caelles_arXiv_2019} dataset that constitutes 60 training videos and 30 validation videos.

\noindent
\textbf{Training} 
While the backbone ResNet(2+1)D layers are initially pre-trained on Kinetics-400, the additional layers are randomly initialized. We use random crops as well as spatial and temporal flips to augment the training data. We, further, generate diverse variations of 32 frame video clip from each video sequence per epoch by dropping a random number of consecutive frames between 1 to 5. The objective function used to optimize the network is discussed in detail in Section \ref{Training Losses}. Additionally, the model is trained using a learning rate of 0.0001 and Adam Optimizer for 100 epochs on Nvidia V100 GPU.

\noindent
\textbf{Inference}
Similar to the training process, whenever we encounter long video sequences during inference, we process the sequence as a 32 frame video (one clip) by dropping intermittent frames. Instance segmentation tags generated from a given clip show well-separated consistent tags per instance. However, in order to associate the given instance throughout the video sequence, we employ interpolation of instance masks in the intermittent frames that were not passed through the inference model.

\subsection{Qualitative Analysis}
\begin{figure}
\begin{center}
\includegraphics[width=1\linewidth]{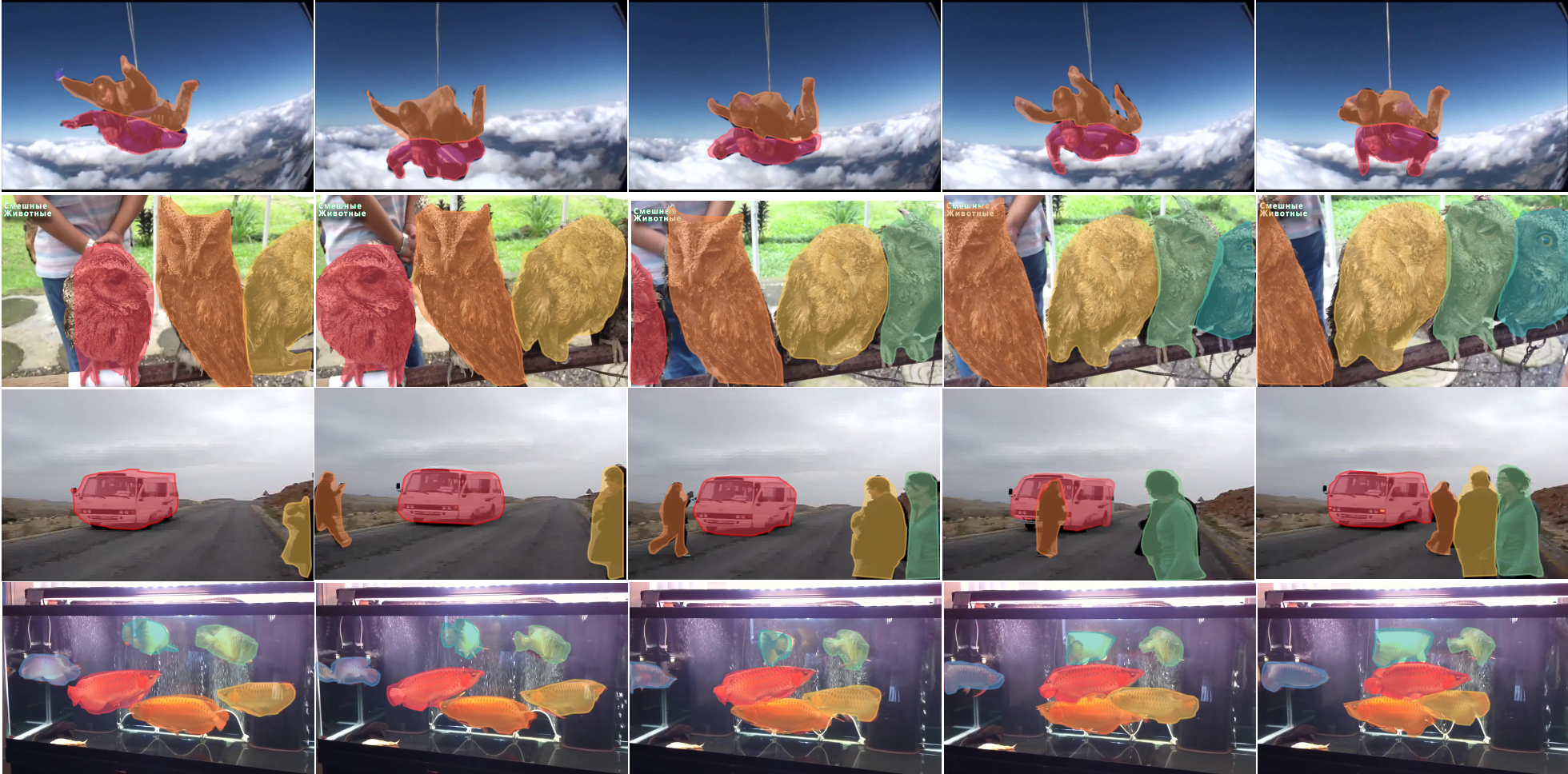}
\end{center}
   \caption{Qualitative results for video instance segmentation on YouTube-VIS validation set}
\label{fig:QualitativeVIS}
\end{figure}

\begin{figure}
\begin{center}
\includegraphics[width=1\linewidth]{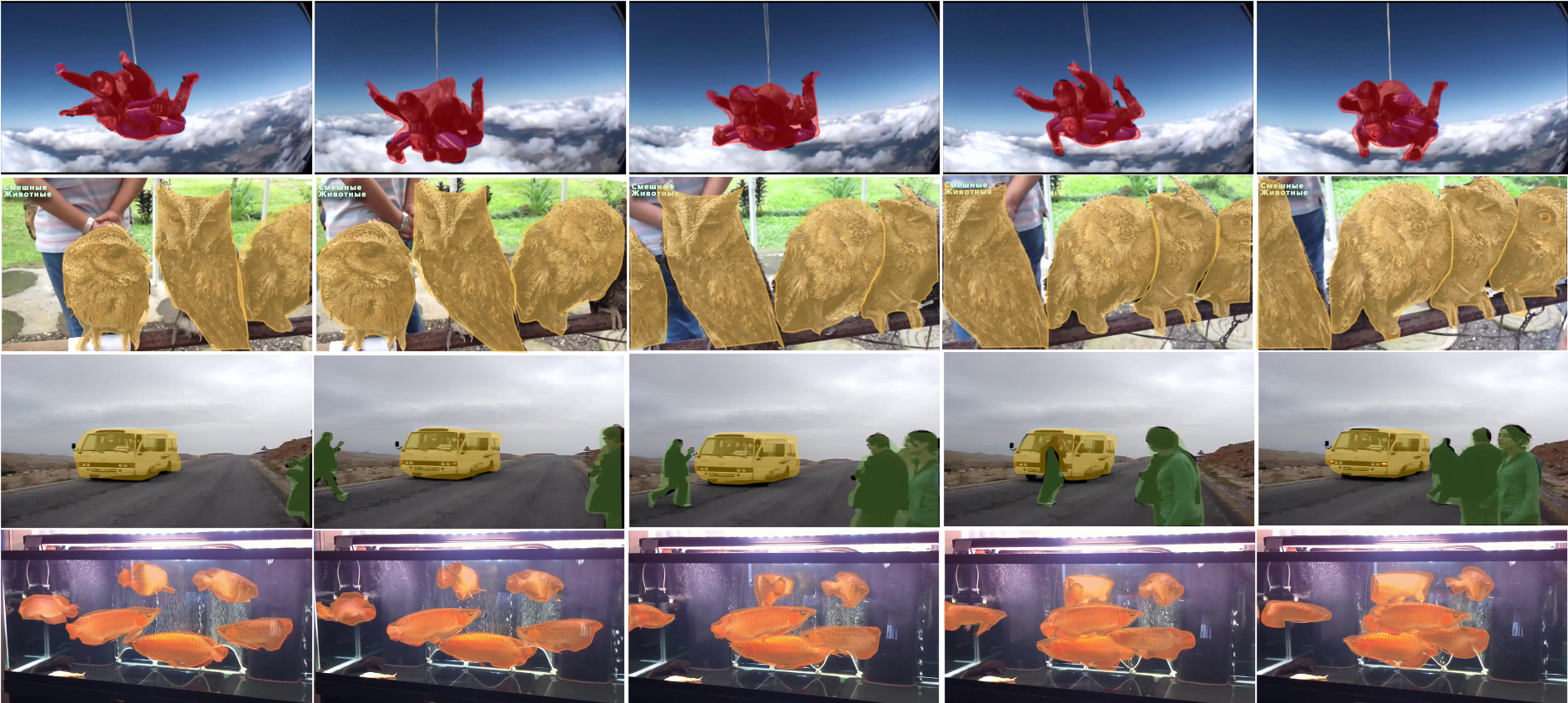}
\end{center}
   \caption{Qualitative results for video semantic segmentation on YouTube-VIS validation set}
\label{fig:QualitativeSS}
\end{figure}

In Figure \ref{fig:QualitativeVIS} , we visualize the VIS results on the Youtube-VIS dataset across four different video clips. Here, different colors have been used to depict distinct object instances based on their predicted tag values. The tag values exhibit consistency for each object instance throughout the video, thus we implicitly associate the given object instance across the sequence. The model depicts its ability to distinguish well between instances belonging to the same object-category. Additionally, our qualitative results suggest that the network is robust to complex variations such as overlapping object constraints, notable object deformation, motion blur, and occlusion. Furthermore, the system predictions appear reasonable even in the case of comparatively smaller objects. Next, we examine the qualitative results for the video semantic segmentation task in Figure \ref{fig:QualitativeSS}. Although we see reasonable class-specific masks generated for the video sequence, we tend to notice that the video semantic segmentation branch suffers due to a lack of customized loss to boost its output.

\subsection{Quantitative Analysis}
Table \ref{table:Youtube-VIS} presents our quantitative results on YouTube-VIS validation set. Here, we observe that unlike other competitive methodologies, our approach is simple with no dependency on external networks and computation-intensive motion cues. 
Similarly, the results on DAVIS’19 Unsupervised validation set in Table \ref{table:Davis} suggest that our end-to-end trained proposal-free network strikes a sound balance between VIS results and processing speed.

\setlength{\tabcolsep}{1.8pt}
\begin{table}[t]
\fontsize{6.5}{8.5}\selectfont
    \begin{center}
        \begin{tabularx}{9.1cm}{@{}l*{9}{c}c@{}}
        \toprule
        {Method} & 
        {Pre-training Data} & 
        {FF} & 
        {P/D} & 
        {\emph{mAP}} &
        {\emph{AP@50}} & 
        {\emph{AP@75}} & 
        {\emph{AR@1}} & 
        {\emph{AR@10}}  \\
        \midrule
        DeepSORT \cite{Wojke2017simple} & Imagenet \cite{russakovsky2015imagenet}, COCO \cite{lin2014microsoft} & &\checkmark & 26.1 & 42.9 & 26.1 & 27.8 & 31.3\\
        FEELVOS \cite{Voigtlaender19CVPR} & Imagenet \cite{russakovsky2015imagenet}, COCO \cite{lin2014microsoft} &\checkmark & & 26.9 & 42.0 & 29.7 & 29.9 & 33.4\\
        OSMN \cite{8578778} & Imagenet \cite{russakovsky2015imagenet}, COCO \cite{lin2014microsoft} & &\checkmark & 27.5 & 45.1 & 29.1 & 28.6 & 33.1\\
        MaskTrack \cite{yang2019video} & Imagenet \cite{russakovsky2015imagenet}, COCO \cite{lin2014microsoft} & &\checkmark & 30.3 & 51.1 & 32.6 & 31.0 & 35.5\\
        STEm-Seg \cite{Athar_Mahadevan20ECCV} & COCO \cite{lin2014microsoft} & & & 34.6 & 55.8 & 37.9 & 34.4 & 41.6\\
        MaskProp \cite{Bertasius_2020_CVPR} & Imagenet \cite{russakovsky2015imagenet}, COCO \cite{lin2014microsoft} & &\checkmark & 46.6 & - & 51.2 & 44.0 & 52.6\\
        Propose-Reduce \cite{Lin_2021_ICCV} & COCO \cite{lin2014microsoft} & &\checkmark & 47.6 & 71.6 & 51.8 & 46.3 & 56.0\\
        \midrule
        {\bf Ours} & Kinetics-400 \cite{kay2017kinetics} & & & {\bf 51.2} & {\bf 74.8} & {\bf 57.9} & {\bf 46.3} & {\bf 57.6}\\
        \bottomrule
    \end{tabularx}
    \end{center}
    \caption{Quantitative results of our method on YouTube-VIS validation set. FF implies first frame proposal, P/D refers to proposals/detections} 
    \label{table:Youtube-VIS}
\end{table}

\setlength{\tabcolsep}{4.3pt}
\begin{table}[t]
\fontsize{6.5}{8.5}\selectfont
    \begin{center}
    \begin{tabularx}{6.7cm}{@{}l*{11}{c}c@{}}
        \toprule
        {Method} & 
        {P/D} & 
        {OF} & 
        {RI} & 
        {\emph{J\&F}} & 
        {\emph{J\textsubscript{Mean}}} & 
        {\emph{F\textsubscript{Mean}}} & 
        {fps}  \\
        \midrule
        RVOS \cite{ventura2019rvos} & & & & 41.2 & 36.8 & 45.7 & 20+ \\
        KIS \cite{cho2019key} &\checkmark & &\checkmark & 59.9 & - & - & - \\
        AGNN \cite{wang2019zero} &\checkmark &\checkmark & & 61.1 & 58.9 & 63.2 & - \\
        STEm-Seg \cite{Athar_Mahadevan20ECCV} & & & & 64.7 & 61.5 & 67.8 & 7\\
        UnOVOST \cite{luiten2020unovost} &\checkmark &\checkmark &\checkmark & 67.0 & 65.6 & 68.4 & \textless{1} \\
        Propose-Reduce \cite{Lin_2021_ICCV} & \checkmark & & & 70.4 & 67.0 & 73.8 & - \\
        \midrule
        {\bf Ours} & & & & {\bf 74.3} & {\bf 72.2} & {\bf 76.4} & {\bf 17}\\
        \bottomrule
    \end{tabularx}
    \end{center}
    \caption{Quantitative results of our method on DAVIS’19 Unsupervised validation set. P/D refers to proposals/detections, OF denotes optical flow, RI implies Re-Id} 
    \label{table:Davis}
\end{table}

\subsection{Ablation Study}
\setlength{\tabcolsep}{4.3pt}
\begin{table}[t]
\fontsize{6.5}{8.5}\selectfont
    \begin{center}
    \begin{tabularx}{6.0cm}{@{}l*{5}{c}c@{}}
        \toprule
         & {\emph{J\&F}} & 
        {\emph{J\textsubscript{Mean}}} & 
        {\emph{F\textsubscript{Mean}}} & \\
        \midrule
        {\bf w/o Losses} & & \\
        \midrule
        {L\textsubscript{spectra}} & 67.2 & 64.4 & 70.0\\
        {L\textsubscript{specter}} & 67.5 & 65.2 & 69.8\\
        {L\textsubscript{tempra}} & 65.8 & 63.3 & 68.2\\
        {L\textsubscript{temper}} & 66.3 & 63.9 & 68.7\\
        \midrule
        {\bf w/o Attention} & & \\
        \midrule
        Spatio-Temporal Attention & 67.9 & 65.4 & 70.3\\
        Tag-Based Attention & 65.9 & 63.5 & 68.3\\
        Self-Attention & 70.3 & 68.9 & 71.6\\
        \midrule
        {\bf w/o Components} & & \\
        \midrule
        Location Embedding & 70.8 & 69.8 & 71.7\\
        Tag-Based Attention + Connector & 64.1 & 62.6 & 65.5\\
        Self-Attention + Connector & 67.4 & 65.0 & 69.7\\
        Tag-Based Attention + Self-Attention + & 61.6 & 60.1 & 63.0\\
        Decoder + Connector & & \\
        \midrule
        {\bf Ours} & {\bf 74.3} & {\bf 72.2} & {\bf 76.4}\\
        \bottomrule
    \end{tabularx}
    \end{center}
    \caption{Ablation study providing performance on DAVIS’19 Unsupervised validation set, when a particular component is eliminated. Here, connector refers to any input or output link corresponding to the module} 
    \label{table:Youtube-VIS-Ablation}
\end{table}

We conduct ablation study using DAVIS’19 Unsupervised dataset and report the analysis in Table \ref{table:Youtube-VIS-Ablation}.  
\\ 
\textbf{Losses} 
To investigate the influence of the spatio-temporal tagging loss on the overall architecture, we test the model performance by eliminating each component of the tagging loss. We notice that eliminating $L_{spectra}$ causes tags of an individual object instance to separate into smaller disjoint clumps, while $L_{specter}$ appears to play a significant role in minimizing tag overlaps amongst distinct object instances. Furthermore, removing $L_{tempra}$ and $L_{temper}$ from the tagging loss impacts the ability of the network to generate consistent tags for a given object instance across the video sequence. 
\\ 
\textbf{Attention} 
Pruning the spatio-temporal attention module results in failure to capture the long-range dependencies across frames.
Next, we eliminate the tag-based attention module, and we see that not only the instance tags generated per frame appear to have poor separation but also the correspondence of instances across the clip seems inconsistent. 
Finally, eliminating the self-attention primarily impacts the semantic segmentation qualitative results, and by the virtue of propagation of loss, it adversely affects the instance segmentation results as well. The self-attention component enables computing enriched frame-based features and provides a comprehensive view of the initial frames to the decoder. 
\\
\textbf{Components} 
Here, we examine the significance of individual and collective components on the model performance. We explicitly add location embeddings in the form of x-y coordinate maps to the ResNet(2+1)D encoded features and based on the outcome we see that it plays a significant role in retaining the positional information. 
In the next experiment, we withdraw the influence of tags on the semantic segmentation decoder by eliminating not just the tag-based attention module but the complete connection that connects tags to the decoder. The drop in the overall network results depicts not only the significance of the tag-based attention but also the importance of loss propagation from the secondary task of video semantic segmentation. Thereafter, we examine the role of initial frame features passed to the semantic segmentation decoder by eliminating the self-attention module along with the connector strapping the input video representation to the semantic segmentation decoder. Here, we notice that the semantic segmentation output is affected significantly, and propagated gradients advertently dampen the instance segmentation results. Ultimately, when we investigate only the instance tag generation branch by removing the video semantic segmentation branch and all the connectors, including the tag-based attention and self-attention modules, we can infer that the secondary task combined with tag-based attention provides a significant boost to the video instance segmentation task.

\section{Conclusion}
In this work, we have introduced a pixel-based novel bottom-up method to solve the VIS problem with minimal network complexities, unlike the alternative top-down approach. Our framework provides a unified strategy to solve the VIS task, and alongside generates decent video semantic segmentation. In the proposed method, we process the video sequence as a single 3D unit to capture the enriched temporal context, and the central idea is based on the concept of generating distinct tags that separate the object instances. The tag-based attention module and tag-based losses present a fresh take on the instance association and separation mechanism. Additionally, experimental evaluations validate that the approach provides competitive results while eliminating redundancy associated with region-proposal based methods. Overall, we explore an alternate direction of research in the area of VIS, and we see the potential to repurpose this approach to address tracking problems as well.

\bibliographystyle{IEEEtran}
\bibliography{IEEEabrv,egbib}

\end{document}